# Bias-Variance Tradeoff in a Sliding Window Implementation of the Stochastic Gradient Algorithm

by

Yakup Ceki Papo

A thesis submitted to Johns Hopkins University

in conformity with the requirements for the degree of

Master of Science in Engineering

Baltimore, Maryland

October, 2019

# Abstract

This paper provides a framework to analyze stochastic gradient algorithms in a mean squared error (MSE) sense using the asymptotic normality result of the stochastic gradient descent (SGD) iterates. We perform this analysis by taking the asymptotic normality result and applying it to the finite iteration case. Specifically, we look at problems where the gradient estimators are biased and have reduced variance and compare the iterates generated by these gradient estimators to the iterates generated by the SGD algorithm. We use the work of Fabian to characterize the mean and the variance of the distribution of the iterates in terms of the bias and the covariance matrix of the gradient estimators. We introduce the sliding window SGD (SW-SGD) algorithm, with its proof of convergence, which incurs a lower MSE than the SGD algorithm on quadratic and convex problems. Lastly, we present some numerical results to show the effectiveness of this framework and the superiority of SW-SGD algorithm over the SGD algorithm.

# Acknowledgments

I'd like to thank my advisor Dr. James Spall for his support of the work presented in this paper and his guidance during times of struggle. It would not have been possible for me to accomplish all this without his mentorship.

I'd like to thank Dr. Avanti Athreya for all the additional ideas and insights that she shared with me throughout the research phase.

I'd like to thank Jingyi Zhu and Long Wang for the valuable advice and opinions that they have shared with me.

I'd like to thank my father for his lifelong mentorship, my mother for her love and my brother for his belief in me. I would not have been able to do any of this without them. Also, a special thanks to my mother for the assistance she has provided in editing this paper.

I'd like to thank all my friends, classmates or otherwise and especially Asya Latifoglu, for their immense support throughout this period.

# Table of Contents

# List of Figures

# Chapter 1

# Introduction

In recent decades, machine learning algorithms and control theory have gained unprecedented importance with the increase of computational power. Many tools from optimization and statistics have played an immense role in the research and development of these algorithms and theories. In particular, stochastic gradient descent (SGD) has been one of the key algorithms used in many machine learning and control tasks.

Stochastic approximation (SA) theory was introduced in the groundbreaking paper of Robbins and Monro (Robbins and Monro, 1951). Originally, Robbins and Monro presented their framework as a solution to the stochastic root-finding problem. Applying their root-finding framework to the gradient of functions to be optimized, gave rise to the SGD algorithm which is used to solve stochastic optimization (SO) problems. Generally a SO problem has the form below:

$$\min_{\boldsymbol{\theta}} \mathbb{E}[Q(\boldsymbol{\theta}, \boldsymbol{V})] \tag{1.1}$$

where $\theta$ is the variable to be optimized and the expectation is with respect to the random variable $V$.

Following (Robbins and Monro, 1951), a new line of research on the SGD algorithm and its variants was set off. A plethora of papers have been published analyzing this algorithm, its variants and their use-cases from both a theoretical and an experimental perspective. Immediately after Robbins and Monro, in (Kiefer and Wolfowitz, 1952) the analysis was extended such that numerical gradients could be used at every iteration of the algorithm to solve problems that are in the form of (1.1). Afterwards, in (Blum, 1954) the author further extended the analysis to the multidimensional $\theta$ regime. Amongst some very popular variants is the momentum algorithm. The momentum algorithm is especially important for our study because the algorithm that we introduce in this paper uses a similar idea to that of momentum. The momentum approach has many different forms. In this paper, we investigate a form of it that is very similar to the one described in (Qian, 1999). Up until today, significant research has been done on trying to improve the performance of these stochastic gradient algorithms and its variants.

Let us take a closer look to the literature that exists on momentum based methods. A widely used version of the momentum algorithm, Nesterov's accelerated gradient (NAG) (Nesterov, 1983), was introduced and analyzed in the deterministic convex optimization setting where the author provided a proof of convergence. Afterwards, NAG and many other versions of the momentum algorithm were analyzed in the SO setting. In (Ruszczynski and Syski, 1983), the authors used averages of gradient estimates in their update and provided a proof of convergence under general noise conditions on the SO problem. As another example in (Spall and Cristion, 1994), the authors use the simultaneous perturbation gradient estimators (introduced in (Spall, 1992)) in a momentum setting (which they refer to as smoothing of the gradient estimator) and illustrate the benefits of the smoothing approach in a neural network (NN) based control problem using numerical studies. They use the simultaneous perturbation gradient estimators to find the optimal parameters of the NN which models the unknown governing equations of the control problem. Additionally, they provide a provide a proof of convergence for the iterates of the algorithm generated by the gradient estimator described above. In more recent years, the development and analysis of the stochastic gradient method with momentum continued especially for training NNs.

Analysis of momentum methods, theoretical and empirical, in NNs and deep learning also exist in more recent works. In (Zhang, Wu, and Zheng, 2006), the authors consider the problem of training a two-layer feedforward NN with a stochastic gradient enhanced with the momentum term. They show weak and strong convergence of the NN weights when the loss function is not necessarily quadratic. Furthermore in (Yan et al., 2018), the authors for the first time provide a proof of convergence of in the non-convex setting where the gradient estimators are constructed as a combination of the Polyak's and Nesterov's momentum in a stochastic setting. In (Sutskever et al., 2013), the authors show how the inferiority of the first-order methods compared to the second-order methods in SO problems can be minimized with a well-designed initialization algorithm and a specific procedure for increasing the momentum term at every iteration. They show the effectiveness of their methods by training a deep auto-encoder with the procedures described above for initialization and momentum. Lastly, (Can, Gurbuzbalaban, and Zhu, 2019) shows that the Wasserstein distance between the $k^{\text{th}}$ iterate of the momentum-based algorithm and the target stationary distribution decays with a linear rate. This result is shown for the strongly-convex function class.

The main reason for this extensive research that has been conducted on SGD and momentum methods is their applicability to many different fields and problems. Almost all machine learning (ML) problems are posed as an optimization problem where the goal is to find the minimizer, $\boldsymbol{\theta}^*$ of some loss function, $L(\boldsymbol{\theta})$. These loss functions are usually complex and finding the minimizer using standard analytical solutions could be difficult and also intractable. In such cases, we use the SGD algorithm to find the minimizer. Many machine learning models such as logistic regression, linear and non-linear regression and neural networks are trained using the SGD algorithm (or its variants).

Due to the complexities of NNs as non-linear functions, applying the SGD algorithm on neural nets gave rise to the well-known backpropagation algorithm (Lecun, 1988). This algorithm provides a way (using the chain rule) to compute the highly complex gradient of the loss function of a neural network with respect to its parameters (i.e. edge weights).

According to the original work of Robbins and Monro, in the standard SGD algorithm, the practitioner is required to use unbiased gradient estimators to find a solution to the problem in (1.1). In our study, we explore problems where using biased gradient estimators with reduced variance is beneficial.

The remainder of this paper is as follows: Chapter 2 is dedicated to reviewing the basic gradient descent algorithm briefly and then introducing the SGD algorithm in detail. Then we continue to Chapter 3 where we introduce the sliding window SGD algorithm with a proof of convergence and asymptotic normality result. In Chapter 4, we provide some numerical studies which give us some concrete understanding of the theoretical results from the previous chapters. Lastly, in Chapter 5, we provide a brief summary of the results and conclusions and put forward some future research directions.

# Chapter 2

# Preliminaries

## 2.1 Gradient Descent

In this paper, we focus on the SGD algorithm. However, before we present the details of SGD, we will start by introducing its deterministic ancestor, gradient descent (GD). The GD algorithm is used to find the minimizer, $\boldsymbol{\theta}^*$, of a differentiable function, $L(\boldsymbol{\theta})$:

$$\boldsymbol{\theta}^* = \arg\min_{\boldsymbol{\theta}} L(\boldsymbol{\theta}). \tag{2.1}$$

At its $k^{th}$ iteration, the GD algorithm performs an update on the current estimate of $\boldsymbol{\theta}^*$, which we denote by $\boldsymbol{\theta}_k$. The update on $\boldsymbol{\theta}_k$ to produce $\boldsymbol{\theta}_{k+1}$ is controlled by the direction of the gradient of $L(\boldsymbol{\theta})$ at $\boldsymbol{\theta}_k$, $\partial L/\partial \boldsymbol{\theta}|_{\boldsymbol{\theta}=\boldsymbol{\theta}_k} = g(\boldsymbol{\theta}_k)$. The iterative algorithm is initiated by at some starting point $\boldsymbol{\theta}_0$. There exist some methods to choose the starting point to ensure faster convergence to $\boldsymbol{\theta}^*$. A widely used method for parameter initialization for neural networks is the Xavier initialization described in (Glorot and Bengio, 2010). In practice, especially for optimizing simpler functions than the loss function of neural networks, random initialization is also used. Overall, the update rule of the GD algorithm has the following form:

$$\boldsymbol{\theta}_{k+1} = \boldsymbol{\theta}_k - a_k \boldsymbol{g}(\boldsymbol{\theta}_k), \quad k = 0, 1, 2 \ldots \tag{2.2}$$

Here $\{a_k\}$ is a sequence of numbers, which we call the gain sequence (also called step size). For the GD algorithm, the terms of the gain sequence are usually chosen to be a constant (i.e. $a_k = a$ for $k = 1, 2, \ldots$), a decaying sequence of positive real numbers (i.e. $\lim_{k \to \infty} a_k = 0$), or a sequence of numbers optimized by the line search procedure (which might be computationally expensive depending on the problem). For the decaying gain sequences, a standard form of $a_k = a/(k+1)^\alpha$ is used for some fixed $a > 0$ and $\alpha \geq 0$. In practice, the practitioner tunes the parameters of the step size to achieve fast convergence. Thorough analysis and convergence proofs of the GD algorithm exist in both the fixed and the decaying gain sequence regimes. Especially, the convergence properties of the GD algorithm to optimize a convex loss function, $L(\boldsymbol{\theta})$ are well known. For example, it is known that the convergence rate of $\boldsymbol{\theta}_k$ to $\boldsymbol{\theta}^*$ is linear (at every step) when the exact line search method is used at every iteration for finding the optimal step size. In the context of descent methods this rate is called linear; however, the term linear here can be misleading as the convergence rate is $O(c^k)$ for $0 < c < 1$. The proof of this result can be found in Section 9.3.1 of (Boyd and Vandenberghe, 2004).

## 2.2 Stochastic Gradient Descent

In this paper, we focus on the SGD algorithm. Like its deterministic ancestor (i.e. GD), SGD is a descent algorithm that is used to optimize loss functions iteratively. However, in many real life scenarios, we do not have access to the explicit form of $L(\boldsymbol{\theta})$ but instead we express $L(\boldsymbol{\theta})$ as the expectation of the noisy process $Q(\boldsymbol{\theta}, \boldsymbol{V})$. Mathematically we write

$$L(\boldsymbol{\theta}) = \mathbb{E}[Q(\boldsymbol{\theta}, \boldsymbol{V})], \tag{2.3}$$

where the expectation is with respect to a random vector $\boldsymbol{V}$.

In real life scenarios, since we do not know the probability density function (pdf) of $\boldsymbol{V}$, we cannot compute $L(\boldsymbol{\theta})$ explicitly and consequently we cannot compute $\boldsymbol{g}(\boldsymbol{\theta})$ that we would use in the GD update. Therefore, the GD algorithm cannot be applied directly to problems where $L$ is expressed as an uncomputable expectation. Nevertheless, our ultimate goal is still to find the minimizer of $L(\boldsymbol{\theta})$, as stated in (2.1). In this case, we apply the SGD algorithm.

We can consider the SGD algorithm as a stochastic analogue of the GD algorithm. At every iteration, instead of updating the parameter vector by a multiple of the true gradient, we update it by an unbiased estimator of the true gradient. We denote the random and unbiased gradient estimator at point $\hat{\boldsymbol{\theta}}_k$ with $\hat{\boldsymbol{g}}(\hat{\boldsymbol{\theta}}_k)$.

For clarity, we use $\boldsymbol{\theta}_k$ and $\hat{\boldsymbol{\theta}}_k$ to denote the parameter vectors in the GD and SGD settings, respectively. Additionally, we assume that $\boldsymbol{\theta}_k$ and $\hat{\boldsymbol{\theta}}_k$ are $p$-dimensional real vectors. The SGD update equation is almost identical to that of GD. The only difference is that in SGD we use the stochastic and unbiased gradient estimate to update the parameter vector, not the true gradient (because we do not know it):

$$\hat{\boldsymbol{\theta}}_{k+1} = \hat{\boldsymbol{\theta}}_k - a_k \hat{\boldsymbol{g}}(\hat{\boldsymbol{\theta}}_k), \ \ k = 0, 1, 2 \ldots \tag{2.4}$$

Once again, we require $\hat{\boldsymbol{g}}(\hat{\boldsymbol{\theta}}_k)$ to be an unbiased estimator of $\boldsymbol{g}(\hat{\boldsymbol{\theta}}_k)$ (i.e. $\mathbb{E}[\hat{\boldsymbol{g}}(\hat{\boldsymbol{\theta}}_k)|\hat{\boldsymbol{\theta}}_k] = \boldsymbol{g}(\hat{\boldsymbol{\theta}}_k)$). Now let us try to express $\boldsymbol{g}(\boldsymbol{\theta})$ as the derivative of the expectation in (2.3). By doing so, we will try to express $\boldsymbol{g}(\boldsymbol{\theta})$ as an expectation of a random vector.

Before getting into the computation of this derivative in (2.5) below, let us note that the the interchange of the derivative and the integral that occurs as we go from the first line to the second line of (2.5) is not a trivial step. Both $Q(\boldsymbol{\theta}, V)$ and $p_V$ in (2.5) have to satisfy a set of conditions for that interchange to be valid. These conditions are given in page 509 of (Spall, 2003).

In our computation, we consider the case where the probability distribution of the noise, $V$, is dependent on $\boldsymbol{\theta}$. Additionally, suppose that the random vector $V$ takes values from the set $\Lambda$. Assuming that the random vector $V$ has an associated pdf, we can express $g(\boldsymbol{\theta})$ as in page 129 of (Spall, 2003):

$$\begin{aligned}
g(\boldsymbol{\theta}) = \frac{\partial L}{\partial \boldsymbol{\theta}} &= \frac{\partial}{\partial \boldsymbol{\theta}} \mathbb{E}[Q(\boldsymbol{\theta}, V)] = \frac{\partial}{\partial \boldsymbol{\theta}} \int_\Lambda Q(\boldsymbol{\theta}, v) p_V(v|\boldsymbol{\theta}) dv \\
&= \int_\Lambda \left[ \frac{\partial Q(\boldsymbol{\theta}, v)}{\partial \boldsymbol{\theta}} p_V(v|\boldsymbol{\theta}) + Q(\boldsymbol{\theta}, v) \frac{\partial p_V(v|\boldsymbol{\theta})}{\partial \boldsymbol{\theta}} \right] dv \\
&= \int_\Lambda \left[ \frac{\partial Q(\boldsymbol{\theta}, v)}{\partial \boldsymbol{\theta}} + Q(\boldsymbol{\theta}, v) \frac{1}{p_V(v|\boldsymbol{\theta})} \frac{\partial p_V(v|\boldsymbol{\theta})}{\partial \boldsymbol{\theta}} \right] p_V(v|\boldsymbol{\theta}) dv \\
&= \int_\Lambda \left[ \frac{\partial Q(\boldsymbol{\theta}, v)}{\partial \boldsymbol{\theta}} + Q(\boldsymbol{\theta}, v) \frac{\partial \log(p_V(v|\boldsymbol{\theta}))}{\partial \boldsymbol{\theta}} \right] p_V(v|\boldsymbol{\theta}) dv \\
&= \mathbb{E} \left[ \frac{\partial Q(\boldsymbol{\theta}, v)}{\partial \boldsymbol{\theta}} + Q(\boldsymbol{\theta}, v) \frac{\partial \log(p_V(v|\boldsymbol{\theta}))}{\partial \boldsymbol{\theta}} \right].
\end{aligned} \quad (2.5)$$

As we can see in expression (2.5), in the most general case of dependency of the pdf of $V$ on $\boldsymbol{\theta}$, one sample of $[\partial Q(\boldsymbol{\theta}, v)/\partial \boldsymbol{\theta} + Q(\boldsymbol{\theta}, v) \partial \log(p_V(v|\boldsymbol{\theta}))/\partial \boldsymbol{\theta}]_{\boldsymbol{\theta}=\hat{\boldsymbol{\theta}}_k}$ can be used as an unbiased estimator of $g(\hat{\boldsymbol{\theta}}_k)$ and can be plugged into the place of $\hat{g}(\hat{\boldsymbol{\theta}}_k)$ in our update equation (2.4).

As we mentioned before, if we do not know the pdf of $V$, obtaining an unbiased estimate of $g(\hat{\boldsymbol{\theta}}_k)$ is not possible in this general case. When the noise distribution is not dependent on $\hat{\boldsymbol{\theta}}_k$, then the unbiased gradient estimate reduces to only $\partial Q(\boldsymbol{\theta}, v)/\partial \boldsymbol{\theta}|_{\boldsymbol{\theta}=\hat{\boldsymbol{\theta}}_k}$ since in this case $\partial \log(p_V(v|\boldsymbol{\theta}))/\partial \boldsymbol{\theta}|_{\boldsymbol{\theta}=\hat{\boldsymbol{\theta}}_k} = 0$.

### 2.2.1 Simple Example to Illustrate Effects of Using Biased Gradient Estimators

In Chapter 3, we introduce our sliding window gradient estimator, which is a biased estimator of the true gradient. In this and the next subsection, we illustrate generically how a biased estimator may be superior to an unbiased estimator in a SO setting.

Let us observe how the convergence of $\hat{\boldsymbol{\theta}}_k$ to $\boldsymbol{\theta}^*$ is affected if the term in the gradient containing $\partial \log(p_V(v|\boldsymbol{\theta}))/\partial \boldsymbol{\theta}$ is ignored in a problem where the noise depends on the parameter vector. To see how the convergence is affected let us consider the problem below:

$$L(\theta) = \mathbb{E}[Q(\theta, V)], \quad Q(\theta, V) = f(\theta, Z) + W, \quad W \sim N(\theta^2, \sigma^2 \theta^2)$$

$$f(\theta, Z) = (\theta - 6)^2 Z, \quad Z \sim N(1, 1), \quad V = (Z, W). \tag{2.6}$$

We are going to assume that $Z$ and $W$ are independent (i.e. $p_V(v) = p_W(w)p_Z(z)$). Using this, we can immediately calculate the true loss function $L(\theta)$ and the true gradient $g(\theta)$ as:

$$L(\theta) = \mathbb{E}[Q(\theta, V)] = \mathbb{E}[f(\theta, Z)] + \mathbb{E}[W] = (\theta - 6)^2 + \theta^2$$

$$L(\theta) = (\theta - 6)^2 + \theta^2 \implies g(\theta) = \frac{\partial L}{\partial \theta} = 4\theta - 12. \tag{2.7}$$

We further know that $\theta^*$ satisfies $g(\theta^*) = 0$. So for this example, $\theta^* = 3$ and $L(\theta^*) = 18$. It is also important to note that $V = (Z, W)$ is dependent on $\theta$ because the pdf of $W$ depends on $\theta$. Therefore, in order to have an unbiased gradient estimate we have to sample from the random variable that is inside the expectation in (2.5). To do so, let us compute the necessary terms required for unbiased gradient estimate calculation:

$$\frac{\partial Q(\theta, v)}{\partial \theta} = \frac{\partial}{\partial \theta}(f(\theta, Z) + W)$$

$$= \frac{\partial f(\theta, Z)}{\partial \theta} = (2\theta - 12)Z. \tag{2.8}$$

Note that $\partial W/\partial \theta = 0$ because as we sample gradient estimates, we condition on a fixed value of the noise (i.e. a fixed value of $W$ and $Z$). Since, the gradient of a constant w.r.t. $\theta$ is zero we get $\partial W/\partial \theta = 0$. Now let's compute $\partial \log(p_V(v|\theta))/\partial \theta$:

$$\frac{\partial \log(p_V(v|\theta))}{\partial \theta} = \frac{\partial \log(p_W(v|\theta))}{\partial \theta} + \frac{\log(p_Z(z))}{\partial \theta} = \frac{\partial \log(p_W(v|\theta))}{\partial \theta}$$

$$= \frac{\partial}{\partial \theta} \log\left(\frac{1}{2\pi\sigma\theta} \exp\left[-\frac{(w-\theta^2)^2}{2\sigma^2\theta^2}\right]\right)$$

$$= \frac{\partial}{\partial \theta}\left(-\log(2\pi\sigma) - \log(\theta) - \frac{(w-\theta^2)^2}{2\sigma^2\theta^2}\right)$$

$$= -\frac{1}{\theta} - \frac{\partial}{\partial \theta}\left(\frac{(w-\theta^2)^2}{2\sigma^2\theta^2}\right)$$

$$= -\frac{1}{\theta} - \frac{\theta^4 - w^2}{\sigma^2\theta^3}. \tag{2.9}$$

It can be confirmed that $\mathbb{E}\left[\partial Q(\theta, V)/\partial \theta + Q(\theta, V)\partial \log(p_V(V|\theta))/\partial \theta\right] = g(\theta)$. Essentially, in this setup $\mathbb{E}[\partial Q/\partial \theta] \neq g(\theta)$ (it is biased). We will define the unbiased and the biased gradient estimates as:

$$\hat{g}_{\text{unbiased}}(\theta) = \frac{\partial Q(\theta, V)}{\partial \theta} + Q(\theta, V)\frac{\partial \log(p_V(V|\theta))}{\partial \theta}$$

$$= (2\theta - 12)Z + (\theta^2 Z + W)(-\frac{1}{\theta} - \frac{\theta^4 - W^2}{\sigma^2\theta^3})$$

$$\hat{g}_{\text{biased}}(\theta) = \frac{\partial Q(\theta, V)}{\partial \theta} = (2\theta - 12)Z. \tag{2.10}$$

Now notice that as we increase $\sigma$, we increase the variance of $W$ which increases the variance of $\hat{g}_{\text{unbiased}}(\theta)$. As we can see from (2.10), since $\hat{g}_{\text{biased}}(\theta)$ does not depend on $\sigma$, changing the value of $\sigma$ has no effect on $\hat{g}_{\text{biased}}(\theta)$. To compare the performances of $\hat{g}_{\text{unbiased}}(\theta)$ and $\hat{g}_{\text{biased}}(\theta)$ as gradient estimators for the SGD algorithm, we run our experiment at different values of $\sigma$ and observe how $\hat{g}_{\text{biased}}(\theta)$ becomes more beneficial to use at high values of $\sigma$ (i.e. when $\hat{g}_{\text{unbiased}}(\theta)$ has high variance).

### 2.2.2 Results of the Simple Example

In this section, we compare the performances of $\hat{g}_{\text{biased}}(\theta)$ and $\hat{g}_{\text{unbiased}}(\theta)$ in finding $\theta^*$ using the SGD update in (2.4) for the problem that we presented in (2.6). We do this comparison by plotting the empirical MSE of the iterates, $\hat{\theta}_k$, as a function of $k$. For all experiments, $\hat{\theta}_0 = 7$ and $\sigma \in \{50, 200, 240, 300\}$. We used a decaying gain sequence of the form $a_k = a/(k+1)^\alpha$ where $\alpha = 0.501$ and $a = 250/\sigma$. For each $\sigma$, we performed a separate SGD run with the gain sequence parameter $a$ tuned to a value that was high enough to guarantee large steps in each update and low enough to prevent overshooting.

Figure 2.1 shows the empirical MSE as a function of $k$ under different values of $\sigma$. The plotted curves were generated by averaging 1000 independent runs. Specifically, let $\hat{\theta}_k^{(j)}$ be the $\hat{\theta}_k$ estimate obtained in the $j^{\text{th}}$ run. For each $k \in \{0, 1, 2 \ldots, 10^4\}$, we calculate $(1/1000) \sum_{j=1}^{1000} (\hat{\theta}_k^{(j)} - \theta^*)^2$ and plot this empirical MSE as a function $k$ for the biased and the unbiased gradient cases.

As we can see in Figure 2.1, the increase in the $\sigma$ values result in slower and noisier convergence for the orange (solid) curves. Slower convergence is because an increase in $\sigma$ leads to a larger variance in the gradient estimators. However, let us notice that in all four plots (even in the $\sigma = 300$ case where

# Empirical MSE of $\hat{\theta}_k$ vs. $k$

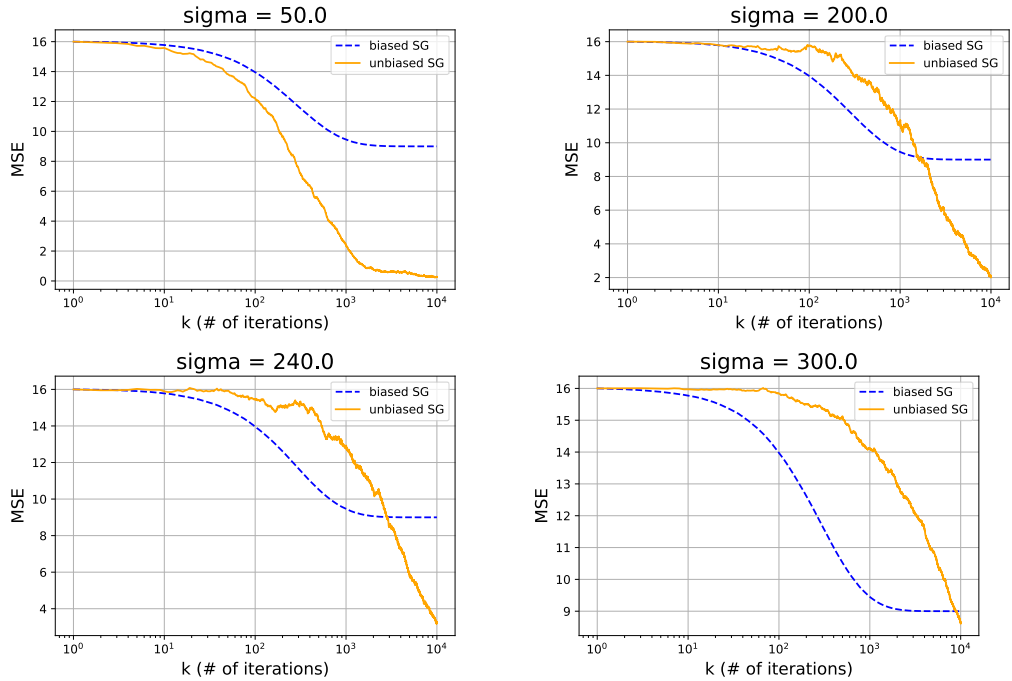

**Figure 2.1:** An illustration of how the level of bias/variance in the gradient estimator affects the convergence behavior of the SGD algorithm. In the plots, the blue (dashed) and the orange (solid) curves represent the empirical MSE of the $\hat{\theta}_k$ sequences that were obtained using the biased and unbiased gradient estimators respectively. In all plots, the blue curves are identical, whereas the orange curves are obtained by increasing the magnitude of the variance in the unbiased gradient estimators. This is done by changing the value of $\sigma$.

the variance of the unbiased gradient estimator is the highest), asymptotically the orange curves outperform the blue (dashed) curves. Another observation is that the blue curves outperform the orange curves for the first $K$ iterations, where $K$ depends on the magnitude of the variance in the unbiased gradient estimator. For instance, if we take a look at the plot for $\sigma = 300$, we will see that the blue curve is outperforming the orange curve for almost all of the first $10^4$ iterations.

It is important to note that in this example the bias of the gradient estimators that gave rise to the blue curves is persistent (i.e. the bias never decays to zero). Therefore, asymptotically, the blue curves will converge to a non-zero MSE value whereas the MSE values of the orange curves will converge to zero as long as the variance of their unbiased gradient estimators is finite. However, as we observe in Figure 2.1, in the non-asymptotic sense, the $\hat{\theta}_k$ sequence generated by biased gradient estimators (blue curves) outperforms the $\hat{\theta}_k$ sequence generated by unbiased gradient estimators (orange curves).

### 2.2.3 Asymptotic Normality of the SGD Iterates

We now establish the asymptotic normality of the SGD iterates. A general form of asymptotic normality was first shown in (Fabian, 1968), where the author was able to completely express the mean and the variance of the asymptotic distribution in terms of the problem parameters. We use the framework constructed by Fabian. The main result is the following:

$$k^{\alpha/2}(\hat{\boldsymbol{\theta}}_k - \boldsymbol{\theta}^*) \xrightarrow{\text{dist}} N(\boldsymbol{\mu}, \boldsymbol{\Sigma}) \qquad (2.11)$$

Using the analysis in (Fabian, 1968), we derive the exact forms of $\boldsymbol{\mu}$ and $\boldsymbol{\Sigma}$ for the SGD algorithm. The $\alpha$ in (2.11) is determined by the decay rate used for the gain sequence and as we will see in the next chapter to guarantee convergence to $\boldsymbol{\theta}^*$, $\alpha$ has to be chosen from a narrow range. Now let us introduce the framework in the Fabian paper adhering to the notation that he used. Before we proceed with Fabian, let us recall that our iterate, $\hat{\boldsymbol{\theta}}_k$, is a $p$-dimensional vector with real-valued entries. In Fabian's framework, we have two random vector sequences $\{V_k\}, \{T_k\}$ where each $V_k, T_k \in \mathbb{R}^p$ and two random matrix sequences $\{\Gamma_k\}, \{\Phi_k\}$ where each $\Gamma_k, \Phi_k \in \mathbb{R}^{p \times p}$. For these sequences, let $V, T, \Gamma, \Phi$ denote the quantities that the sequences $\{V_k\}, \{T_k\}, \{\Gamma_k\}, \{\Phi_k\}$ converge to, respectively. (Let us note that Fabian assumes that $\Gamma$ is a symmetric matrix which can be a restrictive assumption. In (Hernandez and Spall, 2019), the authors generalize the assumption to allow $\Gamma$ to be non-symmetric which is useful in some SA applications.) In what follows, we provide the recursive equation from (Fabian, 1968) that makes use of the random sequences defined above. In the same paper, Fabian was able to show that the iterates of this recursive equation are asymptotically normal provided that $\{V_k\}, \{T_k\}, \{\Gamma_k\}, \{\Phi_k\}$ and $V, T, \Gamma, \Phi$ satisfy the conditions given in Theorem 2.2 of (Fabian, 1968). The recursive equation is given as:

$$(\hat{\boldsymbol{\theta}}_{k+1} - \boldsymbol{\theta}^*) = (\boldsymbol{I} - k^{-\alpha}\boldsymbol{\Gamma}_k)(\hat{\boldsymbol{\theta}}_k - \boldsymbol{\theta}^*) + k^{-(\alpha+\beta)/2}\boldsymbol{\Phi}_k \boldsymbol{V}_k + k^{-\alpha-\beta/2}\boldsymbol{T}_k. \quad (2.12)$$

where $k$ is a positive integer.

More specifically, Fabian showed this asymptotic normality result for the random vector $k^{\beta/2}(\hat{\boldsymbol{\theta}}_k - \boldsymbol{\theta}^*)$. For our purposes, we will give the expressions for these four variables such that once these expressions are plugged into (2.12), the recursive equation will reduce to the simple SGD update. Furthermore, we will show that the expressions we have for $\Gamma_k, \Phi_k, V_k, T_k$ satisfy the necessary conditions to establish the asymptotic normality result. Let us consider the following assignments:

$$\boldsymbol{\Gamma}_k = a\boldsymbol{H}_k, \boldsymbol{\Phi}_k = -a\boldsymbol{I}, \boldsymbol{V}_k = \hat{\boldsymbol{g}}(\hat{\boldsymbol{\theta}}_k) - \mathbb{E}[\hat{\boldsymbol{g}}(\hat{\boldsymbol{\theta}}_k)|\hat{\boldsymbol{\theta}}_k], \boldsymbol{T}_k = \boldsymbol{0}, \beta = \alpha, \quad (2.13)$$

where, as in (Hernandez and Spall, 2019), the $i^{\text{th}}$ row of $H_k$ is equal to the $i^{\text{th}}$ row of the Hessian of $L(\boldsymbol{\theta})$ evaluated at $\boldsymbol{\theta} = (1 - \lambda_i)\hat{\boldsymbol{\theta}}_k + \lambda_i \boldsymbol{\theta}^*$ for some $\lambda_i \in [0,1]$. Let us proceed with showing that the variable assignments we have provided in (2.13) actually give rise to the SGD update when they are plugged into (2.12):

$$(\hat{\boldsymbol{\theta}}_{k+1} - \boldsymbol{\theta}^*) = (I - ak^{-\alpha} H_k)(\hat{\boldsymbol{\theta}}_k - \boldsymbol{\theta}^*)$$

$$- ak^{-(\alpha+\beta)/2}\left[\hat{\boldsymbol{g}}(\hat{\boldsymbol{\theta}}_k) - \mathbb{E}[\hat{\boldsymbol{g}}(\hat{\boldsymbol{\theta}}_k)|\hat{\boldsymbol{\theta}}_k]\right]$$

$$= (\hat{\boldsymbol{\theta}}_k - \boldsymbol{\theta}^*) - ak^{-\alpha} H_k(\hat{\boldsymbol{\theta}}_k - \boldsymbol{\theta}^*)$$

$$- ak^{-\alpha}\hat{\boldsymbol{g}}(\hat{\boldsymbol{\theta}}_k) + ak^{-\alpha}\mathbb{E}[\hat{\boldsymbol{g}}(\hat{\boldsymbol{\theta}}_k)|\hat{\boldsymbol{\theta}}_k]$$

$$= (\hat{\boldsymbol{\theta}}_k - \boldsymbol{\theta}^*) - a_k \hat{\boldsymbol{g}}(\hat{\boldsymbol{\theta}}_k), \tag{2.14}$$

where the last step is valid because $\mathbb{E}[\hat{\boldsymbol{g}}(\hat{\boldsymbol{\theta}}_k)|\hat{\boldsymbol{\theta}}_k] = H_k(\hat{\boldsymbol{\theta}}_k - \boldsymbol{\theta}^*)$. The reasoning behind this equality is given in the first part of the proof of Proposition 2 in (Spall, 1992).

As we can see in (2.14), the assignments provided in (2.13) give rise to the SGD update equation. Now we will show how the variable assignments in (2.13) relate to the mean, $\mu$, and variance, $\Sigma$, of the asymptotic distribution of the SGD iterates:

$$\mu = (\Gamma - \frac{\beta_+}{2} I) T,$$

$$\Sigma = PMP^T, \tag{2.15}$$

where $\Gamma = P \Lambda P^T$ is the SVD of $\Gamma$, $M^{(i,j)} = (P^T \Phi C \Phi^T P)^{(i,j)} (\Lambda^{(i,i)} + \Lambda^{(j,j)} - \beta_+)^{-1}$ is the entry of $M$ in the $i^{\text{th}}$ row and $j^{\text{th}}$ column, $C = \lim_{k\to\infty} \mathbb{E}[V_k V_k^T]$ is the covariance matrix of the gradient estimator and lastly $\beta_+ = \alpha$ if $\alpha = 1$ and $\beta_+ = 0$ if $\alpha \neq 1$.

For the assignments that we have proposed in (2.13) for SGD, we can trivially see that $T = 0$ and $\Phi = -a I$. For $\Gamma$, using the almost sure convergence of the SGD iterates that is shown in (Blum, 1954) (Blum shows convergence for SA problems, which implies convergence for SGD), we can deduce that $\Gamma = \lim_{k \to \infty} \Gamma_k = a \lim_{k \to \infty} H_k = a H^*$ where $H^*$ is equal to the Hessian of $L(\theta)$ evaluated at $\theta = \theta^*$. With these definitions of the new terms, we have fully characterized the mean and the variance of the asymptotic distribution of the SGD iterates.

### 2.2.4 Mean Squared Error (MSE) of the SGD Iterates

One way to measure the performance of stochastic gradient algorithms is by looking at the MSE of the iterates generated by the algorithm. We use the asymptotic normality in Section 2.2.3, to fully characterize the MSE of the iterates. Even though the normality result is valid in an asymptotic sense, we assume that it is also accurate for sufficiently large $k$. Therefore we assume in our analysis that the normality of the iterates, $\hat{\theta}_k$, hold for finite and large $k$. This is an assumption that is widely used in the identification and estimation community, as we can see, for example, in Chapter 4 of (Rubinstein and Kroese, 2016), (Shumway, Olsen, and Levy, 1981), and (Spall and Garner, 1990). For large $k$, we write the normality as,

$$k^{\alpha/2}(\hat{\theta}_k - \theta^*) \sim N(\mu, \Sigma) \implies (\hat{\theta}_k - \theta^*) \sim N\left(\frac{\mu}{k^{\alpha/2}}, \frac{\Sigma}{k^\alpha}\right). \qquad (2.16)$$

Now let us analyze the MSE of $\hat{\boldsymbol{\theta}}_k$ as an estimator of $\boldsymbol{\theta}^*$. Let us note that, as stated in the note after Proposition 2 in (Spall, 1992), the mean and the covariance of the asymptotic distribution match the asymptotic mean and the covariance if $||k^{\alpha/2}(\hat{\boldsymbol{\theta}}_k - \boldsymbol{\theta}^*)||^2$ is uniformly integrable where $||v||$ is the Euclidean norm of a given vector $v$. Assuming that uniform integrability holds and using (2.16), we calculate the MSE as:

$$\mathbb{E}\left[||\hat{\boldsymbol{\theta}}_k - \boldsymbol{\theta}^*||^2\right] = \mathbb{E}\left[||\hat{\boldsymbol{\theta}}_k - \mathbb{E}[\hat{\boldsymbol{\theta}}_k] + \mathbb{E}[\hat{\boldsymbol{\theta}}_k] - \boldsymbol{\theta}^*||^2\right]$$

$$= \mathbb{E}\left[||\hat{\boldsymbol{\theta}}_k - \mathbb{E}[\hat{\boldsymbol{\theta}}_k]||^2 + 2(\hat{\boldsymbol{\theta}}_k - \mathbb{E}[\hat{\boldsymbol{\theta}}_k])^T(\mathbb{E}[\hat{\boldsymbol{\theta}}_k] - \boldsymbol{\theta}^*)\right.$$

$$\left. + ||\mathbb{E}[\hat{\boldsymbol{\theta}}_k] - \boldsymbol{\theta}^*||^2\right]$$

$$= \mathbb{E}\left[||\hat{\boldsymbol{\theta}}_k - \mathbb{E}[\hat{\boldsymbol{\theta}}_k]||^2\right] + 2(\mathbb{E}[\hat{\boldsymbol{\theta}}_k] - \mathbb{E}[\hat{\boldsymbol{\theta}}_k])^T(\mathbb{E}[\hat{\boldsymbol{\theta}}_k] - \boldsymbol{\theta}^*)$$

$$+ \mathbb{E}\left[||\mathbb{E}[\hat{\boldsymbol{\theta}}_k] - \boldsymbol{\theta}^*||^2\right]$$

$$= \mathbb{E}\left[||\hat{\boldsymbol{\theta}}_k - \mathbb{E}[\hat{\boldsymbol{\theta}}_k]||^2\right] + ||\mathbb{E}[\hat{\boldsymbol{\theta}}_k] - \boldsymbol{\theta}^*||^2 = \frac{\text{tr}(\boldsymbol{\Sigma})}{k^\alpha} + \frac{\boldsymbol{\mu}^T\boldsymbol{\mu}}{k^\alpha} \quad (2.17)$$

In (2.17), the $\text{tr}(\boldsymbol{\Sigma})$ and the $\boldsymbol{\mu}^T\boldsymbol{\mu}$ terms are related to the variance and the bias of the gradient estimator respectively. Since SGD converges, we can see that as $k$ grows MSE decays and becomes zero asymptotically.

# Chapter 3

# Bias-Variance Tradeoff in Sliding Window SGD

## 3.1 Sliding Window SGD

In the previous section, we presented a brief summary of the techniques and ideas that are relevant in the theory that we develop and the numerical experiments we conduct. As we saw in Sections 2.2.1 and 2.2.2, using an unbiased gradient estimator does not necessarily give us the best result in terms of the MSE of $\hat{\theta}_k$ in finite samples. Further, as we saw in Figure 2.1, the variance of the gradient estimator used in the update equation also plays an important role in how fast the MSE of $\hat{\theta}_k$ decreases.

We have motivated the potential effectiveness of biased and low variance gradient estimators using an example setup (Section 2.2.1), where the randomness, $V$, in the SO problem depends on the parameter vector, $\theta$. Especially in ML applications and standard learning problems this dependency does not exist. Therefore, the unbiased gradient estimator used in the SGD algorithm in problems where this dependency does not exist simplifies to just $\partial Q(\theta, V)/\partial \theta$.

Let us recall from (2.5) that this simplification is valid since the other term in the unbiased gradient estimator, $Q(\boldsymbol{\theta}, v)\partial\log(p_V(v|\boldsymbol{\theta}))/\partial\boldsymbol{\theta}$, is equal to zero. Nevertheless, using the insights we have obtained from the example in Section 2.2.1, we continue our search to find probable benefits of using biased gradient estimators.

In this section, we construct a new gradient estimator that is biased but has lower variance than the regular SGD gradient estimator. This gradient estimator, which we will call the sliding window (SW) gradient estimator, is a simplified version of the momentum idea that was discussed in Chapter 1. Instead of computing a weighted decaying average over all the past gradient estimates, we will instead look at the uniformly weighted average of the last $t$ gradient estimators. We will call the parameter $t$, the window size. So for a fixed $t$, our SW gradient estimator has the following form:

$$\hat{g}_{\text{SW}}(\hat{\boldsymbol{\theta}}_k, \ldots, \hat{\boldsymbol{\theta}}_{k-t+1}) = \frac{1}{t}\sum_{i=0}^{t-1}\frac{\partial Q(\boldsymbol{\theta}, V_{k-i})}{\partial \boldsymbol{\theta}}\bigg|_{\boldsymbol{\theta}=\hat{\boldsymbol{\theta}}_{k-i}} = \frac{1}{t}\sum_{i=0}^{t-1}\hat{g}(\hat{\boldsymbol{\theta}}_{k-i}) \quad (3.1)$$

Let us now consider this gradient estimator as a potential candidate that might outperform the standard unbiased gradient estimator. Similar to the standard SGD update rule in equation (2.4), at the $k^{th}$ iteration we use the SW gradient estimator to do the update as:

$$\hat{\boldsymbol{\theta}}_{k+1} = \hat{\boldsymbol{\theta}}_k - a_k\hat{g}_{\text{SW}}(\hat{\boldsymbol{\theta}}_k, \hat{\boldsymbol{\theta}}_{k-1}, \ldots, \hat{\boldsymbol{\theta}}_{k-t+1}) \quad (3.2)$$

We call the algorithm that uses the update rule in (3.2) the SW-SGD algorithm. In the following section, we provide a proof of convergence for the SW-SGD algorithm under some conditions.

## 3.2 Proof of Almost Sure Convergence of the SW-SGD Algorithm

Let us start by considering the case where $t = 2$ for the window size. For this case, the SW gradient estimator is simply the average of the current and the previous gradient estimators:

$$\hat{g}_{\text{SW}}(\hat{\boldsymbol{\theta}}_k, \hat{\boldsymbol{\theta}}_{k-1}) = \frac{1}{2}\left[\hat{g}(\hat{\boldsymbol{\theta}}_k) + \hat{g}(\hat{\boldsymbol{\theta}}_{k-1})\right] \tag{3.3}$$

In addition to the conditions that we will present to ensure convergence of the $\hat{\boldsymbol{\theta}}_k$ sequence generated by using the SW gradient estimator in (3.3) to $\boldsymbol{\theta}^*$, we also introduce the assumptions that we impose for our proofs to be valid. These assumptions are:

a. The randomness, $\{V_k\}$, for the given problem is bounded.

b. The loss function is a quadratic convex with a Lipschitz continuous gradient.

c. The noisy loss function has the form: $Q(\boldsymbol{\theta}, V) = VL(\boldsymbol{\theta})$ where $\mathbb{E}[V] = 1$.

d. $\text{Var}(V_k)$ decays at a rate of $O(1/k^{2\gamma})$ where $\gamma \geq 1/2$. We can express the randomness as: $V_k = 1 + X_k/k^\gamma$ where $X_k$ is a mean zero random variable.

Let us recall the definition of a Lipschitz continuous function. A function $f : \mathbb{R}^n \to \mathbb{R}$ is said to be Lipschitz continuous if there exists a real number $M > 0$ (which is called the Lipschitz constant) such that:

$$||f(\boldsymbol{x}) - f(\boldsymbol{y})|| \leq M||\boldsymbol{x} - \boldsymbol{y}|| \ \forall \ \boldsymbol{x}, \boldsymbol{y} \in \text{dom}(f). \tag{3.4}$$

Due to Assumption (a), we cannot use unbounded random vectors for the $\{V_k\}$ sequence (For example, we cannot use Gaussian random vectors, instead we can use truncated Gaussian random vectors as a substitute). Also, we need Assumption (d) for our convergence proof to be valid. A real-life application where this assumption is satisfied is when we average past gradient estimators to find the current gradient estimator. In Chapter 5 of (Kushner and Yin, 2003), the authors provide a set of conditions that are required to be satisfied in order to guarantee convergence. This set of conditions may be modified in certain cases, which gives rise to a weaker modes of convergence. In our analysis, we will show convergence of the $\hat{\theta}_k$ sequence in an almost sure sense (i.e. the strongest standard mode of convergence).

### 3.2.1 Conditions for Almost Sure Convergence

As stated in page 126 of (Kushner and Yin, 2003), we now present the conditions that needed to be satisfied to ensure convergence. Before the conditions, we define the bias, $\beta_k$, in $\hat{g}_{SW}(\hat{\theta}_k, \hat{\theta}_{k-1})$ with the equation $\mathbb{E}[\hat{g}_{SW}(\hat{\theta}_k, \hat{\theta}_{k-1})|\hat{\theta}_i, \forall i \leq k] = g(\hat{\theta}_k) + \beta_k$. Following the condition statements below, we continue with showing that the SW gradient estimator satisfies all these conditions below:

1. $\sup_k \mathbb{E}[||\hat{g}_{SW}(\hat{\theta}_k, \hat{\theta}_{k-1})||] < \infty$,

2. $g(\cdot)$ is continuous,

3. $\sum_{i=0}^{\infty} a_i^2 < \infty$, $\sum_{i=0}^{\infty} a_i = \infty$,

4. $\sum_{i=0}^{\infty} a_i ||\beta_i|| < \infty$ a.s.

Let us recall that as we have defined previously, the sequence $\{a_i\}$ is the decaying gain sequence used in the update equations where each $a_i$ usually has the form $a/(i+1)^\alpha$ for some fixed $a$ and $\alpha$.

In our analysis, the candidate gradient estimator in our update equation is $\hat{g}_{SW}(\hat{\theta}_k, \hat{\theta}_{k-1})$. In other words, we will show that the four conditions above are satisfied for $\hat{g}_{SW}(\hat{\theta}_k, \hat{\theta}_{k-1})$. Before we get into confirming that all the conditions are satisfied, let us compute the bias, $\beta_k$, of our gradient estimator:

$$\beta_k = \mathbb{E}[\hat{g}_{SW}(\hat{\theta}_k, \hat{\theta}_{k-1}) - g(\hat{\theta}_k) | \hat{\theta}_i, \forall i \leq k]$$

$$= \mathbb{E}\left[\frac{1}{2}[\hat{g}(\hat{\theta}_k) + \hat{g}(\hat{\theta}_{k-1})] - g(\hat{\theta}_k) \bigg| \hat{\theta}_i, i \leq k\right]$$

$$= \frac{1}{2}\mathbb{E}\left[\hat{g}(\hat{\theta}_k) | \hat{\theta}_k\right] + \frac{1}{2}\mathbb{E}\left[\hat{g}(\hat{\theta}_{k-1}) | \hat{\theta}_k, \hat{\theta}_{k-1}\right] - g(\hat{\theta}_k)$$

$$= -\frac{g(\hat{\theta}_k)}{2} + \frac{\hat{g}(\hat{\theta}_{k-1})}{2} - \left(\frac{g(\hat{\theta}_{k-1})}{2} - \frac{g(\hat{\theta}_{k-1})}{2}\right)$$

$$= \frac{g(\hat{\theta}_{k-1}) - g(\hat{\theta}_k)}{2} + \frac{\hat{g}(\hat{\theta}_{k-1}) - g(\hat{\theta}_{k-1})}{2}$$

$$= \frac{g(\hat{\theta}_{k-1}) - g(\hat{\theta}_k)}{2} + \frac{g(\hat{\theta}_{k-1})(V_{k-1} - 1)}{2}. \tag{3.5}$$

Now that all the variables are defined, we will continue with showing that all four conditions for convergence that we mentioned above are satisfied for $\hat{g}_{SW}(\hat{\theta}_k, \hat{\theta}_{k-1})$. Let us start by considering Condition 1:

$$\sup_k \mathbb{E}[||\hat{g}_{SW}(\theta)||] = \sup_k \mathbb{E}\left[\left\|\frac{1}{2}[\hat{g}(\hat{\theta}_k) + \hat{g}(\hat{\theta}_{k-1})]\right\|\right]$$

$$\leq \sup_k \mathbb{E}\left[||\hat{g}(\hat{\theta}_k)||\right] + \mathbb{E}\left[||\hat{g}(\hat{\theta}_{k-1})||\right]. \tag{3.6}$$

We can infer from the inequality in (3.6) that as long as the unbiased gradient estimators that are used in the classical SGD algorithm are finite in expectation, our SW gradient estimator (which is derived from the SGD gradient estimators) will be too. Because we are considering problems with bounded randomness in our analysis, this condition is satisfied for the SGD gradient estimators and therefore for the SW-SGD estimators. It might also be possible for $||\hat{\boldsymbol{\theta}}_k|| \to \infty$ which might cause $\mathbb{E}[||\hat{g}_{\text{SW}}(\hat{\boldsymbol{\theta}}_k, \hat{\boldsymbol{\theta}}_{k-1})||] \to \infty$ violating Condition 1. Fortunately, as stated in (Spall, 1992), this is unexpected to happen in most applications, so we do not consider the event of $||\hat{\boldsymbol{\theta}}_k|| \to \infty$ to be a likely and a restrictive one.

In Condition 2, we require the true gradient of the loss function of the problem to be continuous. This is easily satisfied because in our analysis we study convex and quadratic loss functions, which have Lipschitz continuous gradients. Now, let us consider the general form of quadratic and convex loss functions, $L(\boldsymbol{\theta}) = \boldsymbol{\theta}^T A \boldsymbol{\theta}$ for a positive semi-definite (PSD) $A$. The gradient of this loss function is calculated as:

$$\frac{\partial L}{\partial \boldsymbol{\theta}} = g(\boldsymbol{\theta}) = 2A\boldsymbol{\theta}. \tag{3.7}$$

For any $\boldsymbol{\theta}_1, \boldsymbol{\theta}_2 \in \text{dom}(L)$ we can say the following:

$$||2A\boldsymbol{\theta}_1 - 2A\boldsymbol{\theta}_2|| = ||2A(\boldsymbol{\theta}_1 - \boldsymbol{\theta}_2)|| \leq 2 \cdot ||A|| \cdot ||\boldsymbol{\theta}_1 - \boldsymbol{\theta}_2||, \tag{3.8}$$

where $||A||$ is the spectral norm of the matrix $A$.

By letting the Lipschitz constant of $g(\boldsymbol{\theta})$ be $M = 2||A||$, we can observe that the gradient of $L$ is indeed Lipschitz continuous. Since we have established

Lipschitz continuity of the gradient, $g$, we showed that Condition 2 holds.

Condition 3 is about the step sizes that we use in our SW-SGD algorithm. As we have mentioned before, at iteration $i$, we use step size $a_i = a/(i+1)^\alpha$. This form of $a_i$ together with Condition 3, results in $\alpha \in (0.5, 1]$ to be the only acceptable range for $\alpha$. This range for $\alpha$ is also provided in page 109 of (Spall, 2003). Lastly, let us show that the SW gradient estimator also satisfies Condition 4. We start by finding an upper bound for $\sum_{i=0}^{\infty} a_i ||\boldsymbol{\beta}_i||$ as:

$$\sum_i a_i ||\boldsymbol{\beta}_i|| = \sum_i a_i \left\| \frac{g(\hat{\boldsymbol{\theta}}_{i-1}) - g(\hat{\boldsymbol{\theta}}_i)}{2} + \frac{g(\hat{\boldsymbol{\theta}}_{i-1})(V_{i-1} - 1)}{2} \right\|$$

$$\leq \frac{L}{2} \sum_i a_i ||\hat{\boldsymbol{\theta}}_{i-1} - \hat{\boldsymbol{\theta}}_i|| + \sum_i a_i \left\| \frac{g(\hat{\boldsymbol{\theta}}_{i-1})(V_{i-1} - 1)}{2} \right\|$$

$$= \frac{L}{2} \sum_i a_i ||a_{i-1} \hat{g}_{SW}(\hat{\boldsymbol{\theta}}_{i-1}, \hat{\boldsymbol{\theta}}_{i-2})|| + \sum_i a_i \left\| \frac{g(\hat{\boldsymbol{\theta}}_{i-1})(V_{i-1} - 1)}{2} \right\|$$

$$\leq \frac{L}{2} \sum_i a_{i-1}^2 ||\hat{g}_{SW}(\hat{\boldsymbol{\theta}}_{i-1}, \hat{\boldsymbol{\theta}}_{i-2})|| + \sum_i a_i \left\| \frac{g(\hat{\boldsymbol{\theta}}_{i-1})(V_{i-1} - 1)}{2} \right\|$$

$$= \frac{L}{2} \sum_i a_{i-1}^2 \left\| \frac{\hat{g}(\hat{\boldsymbol{\theta}}_{i-1}) + \hat{g}(\hat{\boldsymbol{\theta}}_{i-2})}{2} \right\| + \frac{1}{2} \sum_i a_i \left\| \frac{X_{i-1}}{i^\gamma} \right\| ||g(\hat{\boldsymbol{\theta}}_{i-1})||$$

$$\leq \frac{L}{4} \sum_i a_{i-1}^2 (||\hat{g}(\hat{\boldsymbol{\theta}}_{i-1})|| + ||\hat{g}(\hat{\boldsymbol{\theta}}_{i-2})||) + \frac{1}{2} \sum_i a_i \left\| \frac{X_{i-1}}{i^\gamma} \right\| ||g(\hat{\boldsymbol{\theta}}_{i-1})||$$

$$= \frac{L}{4} \sum_i \frac{a}{i^{2\alpha}} (||\hat{g}(\hat{\boldsymbol{\theta}}_{i-1})|| + ||\hat{g}(\hat{\boldsymbol{\theta}}_{i-2})||)$$

$$+ \frac{1}{2} \sum_i \frac{a}{(i+1)^\alpha} \cdot \frac{1}{i^\gamma} \cdot |X_{i-1}| \cdot ||g(\hat{\boldsymbol{\theta}}_{i-1})|| \tag{3.9}$$

To satisfy Condition 4, we now have to show that the expression we have obtained in (3.9) is bounded almost surely. So mathematically, we would like to show that for an arbitrary sample point $\omega$:

$$P\Bigg(\bigg\{\omega : \frac{L}{4}\sum_i \left(\frac{a}{i}\right)^{2\alpha}(||\hat{g}(\hat{\boldsymbol{\theta}}_{i-1})|| + ||\hat{g}(\hat{\boldsymbol{\theta}}_{i-2})||$$

$$+ \frac{1}{2}\sum_i \frac{a}{(i+1)^\alpha}\cdot\frac{1}{i^\gamma}\cdot|X_{i-1}|\cdot||g(\hat{\boldsymbol{\theta}}_{i-1})||) < C(\omega)\bigg\}\Bigg) = 1$$

$$\text{s.t. } C(\omega) < \infty\ \forall\ \omega. \tag{3.10}$$

In our study, we are considering stochastic optimization problems that have bounded noise. Therefore, the noisy gradients (i.e. $\hat{g}$) are always perturbed a finite amount and we know that the noisy gradients themselves are bounded (i.e. $\hat{g}(\hat{\boldsymbol{\theta}}) < K\ \forall\ \hat{\boldsymbol{\theta}}$ for some finite $K$). So all we have to do now is to show that $\sum_i a/i^{2\alpha} < \infty$. This is clearly true because due to Condition 3. In this condition, we require that $\alpha \in (0.5, 1]$, so $2\alpha \in (1, 2]$ and this is sufficient for the first summation in (3.10) to converge. For the second summation, we have that $|X_{i-1}|$ and $||g(\hat{\boldsymbol{\theta}}_{i-1})||$ are bounded due to the fact that $g(\hat{\boldsymbol{\theta}}_{i-1})$ is a linear function of $\hat{\boldsymbol{\theta}}_{i-1}$. As have mentioned before and as explained in (Spall, 1992), $||\hat{\boldsymbol{\theta}}_{i-1}|| \to \infty$ is a very unlikely event and not expected to happen in most applications which makes $||g(\hat{\boldsymbol{\theta}}_{i-1})|| \to \infty$ an unlikely event in most applications as well. Finally, we know by Condition 3 that $\alpha \in (0.5, 1]$ and by Assumption (d), we know that $\gamma \geq 0.5$. So in the second summation we have two terms that are bounded multiplied with two decaying term which have a total decay rate of $\alpha + \gamma > 1$. Due to these reasons, the second summation converges as well. Hence, (3.10) holds.

Because we have proved the validity of the four conditions that we have provided in the beginning of this section, we know that SW-SGD algorithm converges to the minimizer of the loss function in an almost sure sense. We will be using this result in the following section.

## 3.3 Asymptotic Normality of the SW-SGD Iterates

In this section, we will establish an asymptotic normality result for the SW-SGD iterates similar to how we did it for the SGD iterates. As in Section 2.2.3, we will be using the results from (Fabian, 1968). Additionally, we will be using (Spall, 1992) as a guide to come up with the assignments of $\boldsymbol{\Gamma}_k, \boldsymbol{\Phi}_k, \boldsymbol{V}_k, \boldsymbol{T}_k$ for our SW-SGD algorithm. Let us now propose potential assignments for these four variables and show that when these assignments are plugged into (2.12) the simplified recursive equation is the update for the SW-SGD algorithm.

$$\boldsymbol{\Gamma}_k = a\boldsymbol{H}_k,$$

$$\boldsymbol{\Phi}_k = -a\boldsymbol{I},$$

$$\boldsymbol{V}_k = \hat{\boldsymbol{g}}_{\text{SW}}(\hat{\boldsymbol{\theta}}_k, \hat{\boldsymbol{\theta}}_{k-1}) - \mathbb{E}[\hat{\boldsymbol{g}}_{\text{SW}}(\hat{\boldsymbol{\theta}}_k, \hat{\boldsymbol{\theta}}_{k-1})|\hat{\boldsymbol{\theta}}_k, \hat{\boldsymbol{\theta}}_{k-1}],$$

$$\boldsymbol{T}_k = -ak^{\beta/2}\boldsymbol{\beta}_k,$$

$$\beta = \alpha, \tag{3.11}$$

where $\boldsymbol{H}_k$ is defined as in (2.13).

Now we will proceed by showing that the assignments in (3.11) generate the update equation we have for SW-SGD in (3.2). We will show this by plugging in these assignments into the recursive equation (2.12):

$$(\hat{\boldsymbol{\theta}}_{k+1} - \boldsymbol{\theta}^*) = (\boldsymbol{I} - ak^{-\alpha}\boldsymbol{H}_k)(\hat{\boldsymbol{\theta}}_k - \boldsymbol{\theta}^*)$$

$$- ak^{-(\alpha+\beta)/2}\left[\hat{\boldsymbol{g}}_{SW}(\hat{\boldsymbol{\theta}}_k, \hat{\boldsymbol{\theta}}_{k-1}) - \mathbb{E}[\hat{\boldsymbol{g}}_{SW}(\hat{\boldsymbol{\theta}}_k, \hat{\boldsymbol{\theta}}_{k-1})|\hat{\boldsymbol{\theta}}_k, \hat{\boldsymbol{\theta}}_{k-1}]\right]$$

$$- ak^{-\alpha-\beta/2}k^{\beta/2}\boldsymbol{\beta}_k$$

$$= (\hat{\boldsymbol{\theta}}_k - \boldsymbol{\theta}^*) - ak^{-\alpha}\boldsymbol{H}_k(\hat{\boldsymbol{\theta}}_k - \boldsymbol{\theta}^*) - ak^{-\alpha}\hat{\boldsymbol{g}}_{SW}(\hat{\boldsymbol{\theta}}_k, \hat{\boldsymbol{\theta}}_{k-1})$$

$$+ ak^{-\alpha}\mathbb{E}[\hat{\boldsymbol{g}}_{SW}(\hat{\boldsymbol{\theta}}_k, \hat{\boldsymbol{\theta}}_{k-1})|\hat{\boldsymbol{\theta}}_k, \hat{\boldsymbol{\theta}}_{k-1}] - ak^{-\alpha}\boldsymbol{\beta}_k$$

$$= (\hat{\boldsymbol{\theta}}_k - \boldsymbol{\theta}^*) - a_k\hat{\boldsymbol{g}}_{SW}(\hat{\boldsymbol{\theta}}_k, \hat{\boldsymbol{\theta}}_{k-1}), \tag{3.12}$$

where $k$ is a positive integer and $\mathbb{E}[\hat{\boldsymbol{g}}_{SW}(\hat{\boldsymbol{\theta}}_k, \hat{\boldsymbol{\theta}}_{k-1})|\hat{\boldsymbol{\theta}}_k, \hat{\boldsymbol{\theta}}_{k-1}] = \boldsymbol{H}_k(\hat{\boldsymbol{\theta}}_k - \boldsymbol{\theta}^*) + \boldsymbol{\beta}_k$. The reasoning behind this equality is given in the first part of the proof of Proposition 2 in (Spall, 1992).

As in (2.15), the mean and variance of the asymptotic distribution of $k^{\alpha/2}(\hat{\boldsymbol{\theta}}_k - \boldsymbol{\theta}^*)$ is expressed in terms of $\boldsymbol{\Gamma}, \boldsymbol{\Phi}, \boldsymbol{T}$. For the SW-SGD algorithm, $\boldsymbol{\Gamma} = a\boldsymbol{H}^*$ and $\boldsymbol{\Phi} = -a\boldsymbol{I}$ as in the SGD case (the analysis is the same as in (2.15)). In this case, $\boldsymbol{T}$ is again equal to zero because the bias sequence $\boldsymbol{\beta}_k$ converges to the zero vector. The bias $\boldsymbol{\beta}_k$ converges to the zero vector because, as proven in the previous subsection, the SW-SGD algorithm converges, which makes $\boldsymbol{g}(\hat{\boldsymbol{\theta}}_{k-1}) - \boldsymbol{g}(\hat{\boldsymbol{\theta}}_k)$ converge to the zero vector and $\boldsymbol{g}(\hat{\boldsymbol{\theta}}_{k-1})$ in $\boldsymbol{D}_k$ to converge to the zero vector.

# Chapter 4

# Numerical Experiments

In this chapter we present two numerical experiments to show how the bias-variance tradeoff of the gradient estimators affect the asymptotic MSE of the SGD and SW-SGD iterates.

## 4.1 Simple SGD Experiment

First let us start with a simple problem to show how we use the asymptotic normality result to calculate the MSE of the iterates. Consider the following problem setup:

$$Q(\theta, V) = (\theta - d)^T(\theta - d)V \text{ where } V \sim N(1, r)$$

$$L(\theta) = \mathbb{E}[Q(\theta, V)] = (\theta - d)^T(\theta - d)$$

$$g(\theta) = 2(\theta - d) \tag{4.1}$$

Now let us introduce the two gradient estimators that we use to find $\theta^*$ of the problem in (4.1). One of these gradient estimators is biased with low variance and the other is unbiased with higher variance. These gradient estimators are defined as:

$$\hat{g}_{\text{unbiased}}(\boldsymbol{\theta}) = \frac{\partial Q}{\partial \boldsymbol{\theta}} = 2(\boldsymbol{\theta} - \boldsymbol{a})V,$$

$$\hat{g}_{\text{biased}}(\boldsymbol{\theta}) = \frac{\partial Q}{\partial \boldsymbol{\theta}} + \boldsymbol{b} = 2(\boldsymbol{\theta} - \boldsymbol{a})V + \boldsymbol{b}. \tag{4.2}$$

We control the bias and the variance of the gradient estimators with parameters $\boldsymbol{b}$ for bias and $r_u, r_b$ for the variance of unbiased and biased gradient estimators respectively. We express the distributions of these two gradient estimators as:

$$2(\boldsymbol{\theta} - \boldsymbol{a})V \sim N(2(\boldsymbol{\theta} - \boldsymbol{a}), 4r_u(\boldsymbol{\theta} - \boldsymbol{a})(\boldsymbol{\theta} - \boldsymbol{a})^T)$$

$$2(\boldsymbol{\theta} - \boldsymbol{a})V + \boldsymbol{b} \sim N(2(\boldsymbol{\theta} - \boldsymbol{a}) + \boldsymbol{b}, 4r_b(\boldsymbol{\theta} - \boldsymbol{a})(\boldsymbol{\theta} - \boldsymbol{a})^T) \tag{4.3}$$

We conduct four different experiments where two distinct iterate sequences are produced using the SGD update in (2.4). One of these iterate sequences is produced using the biased and the other is produced using the unbiased gradient estimator in their update. In all four experiments, the vector $\boldsymbol{b}$ is set to the vector of ones, which is the amount of bias in $\hat{g}_{\text{biased}}(\boldsymbol{\theta})$. As for the variances of these gradient estimators, $r_b = 10$, and $r_u \in \{50, 100, 500, 1000\}$ in the four experiments conducted.

In this experiment, since the bias of $\hat{g}_{\text{biased}}(\boldsymbol{\theta})$ is persistent (i.e. bias is non-zero asymptotically), the iterates produced by this gradient estimator do not converge precisely to $\boldsymbol{\theta}^*$. We can also see this from the convergence of the blue (dashed) curves to non-zero values in Figure 4.1. Let us assume that the iterates produced by using the biased gradient converge and let us call the vector that they converge to $\boldsymbol{\theta}^{*\prime}$. Let us now define $w = \boldsymbol{\theta}^* - \boldsymbol{\theta}^{*\prime}$ for later use in calculation of MSEs. Further, let us define $\{\hat{\boldsymbol{\theta}}_k\}$ and $\{\hat{\boldsymbol{\theta}}_k^\prime\}$ as the iterate sequences generated by using the unbiased and biased gradient estimators respectively. Furthermore, let us recall Section 2.2.3 where we derived the expressions for the mean and variance of the distribution of the iterates. We can write the distribution of $\hat{\boldsymbol{\theta}}_k$ as:

$$(\hat{\boldsymbol{\theta}}_k - \boldsymbol{\theta}^*) \sim N\left(\mathbf{0}, \frac{\Sigma}{k^\alpha}\right) \tag{4.4}$$

where $\Sigma$ is the covariance matrix of the asymptotic distribution of $\hat{\boldsymbol{\theta}}_k$ defined as in (2.15).

Similarly, we express the distribution of $\hat{\boldsymbol{\theta}}_k^\prime$ as:

$$(\hat{\boldsymbol{\theta}}_k^\prime - \boldsymbol{\theta}^{*\prime}) \sim N\left(\mathbf{0}, \frac{\Sigma^\prime}{k^\alpha}\right) \implies (\hat{\boldsymbol{\theta}}_k^\prime - \boldsymbol{\theta}^* - w) \sim N\left(\mathbf{0}, \frac{\Sigma^\prime}{k^\alpha}\right)$$

$$\implies (\hat{\boldsymbol{\theta}}_k^\prime - \boldsymbol{\theta}^*) \sim N\left(w, \frac{\Sigma^\prime}{k^\alpha}\right) \tag{4.5}$$

where $\Sigma$ is the covariance matrix of the asymptotic distribution of $\hat{\boldsymbol{\theta}}_k^\prime$ defined as in (2.15).

In our experiments we track empirical MSE of the iterates, $\hat{\boldsymbol{\theta}}_k$, as a function of $k$ and observe its decay. We calculate the empirical MSE in the same way as we did for the simple experiment presented in (2.6). This time we average the MSE values over 10000 independent runs. For each of the four rounds of the experiment, we generate a plot of $\mathbb{E}[||\hat{\boldsymbol{\theta}}_k - \boldsymbol{\theta}^*||^2]$ vs. $k$. On each of these plots, we also show a green vertical line representing the intersection point of the theoretical MSE curves. We then compare the theoretical intersection to the empirical one that we observe from the intersection of the blue (dashed) and the orange (solid) curves. We calculate this theoretical intersection by using the parameters defined in (2.15) for the asymptotic mean and variance of the iterates.

More specifically, the iteration count that the vertical green line corresponds to is calculated by finding the value of $k$ that satisfies $\mathbb{E}[||\hat{\boldsymbol{\theta}}_k - \boldsymbol{\theta}^*||^2] = \mathbb{E}[||\hat{\boldsymbol{\theta}}'_k - \boldsymbol{\theta}^*||^2]$. In this case, using (4.4) and (4.5), the theoretical intersection point is the value of $k$ that satisfies:

$$\mathbb{E}[||\hat{\boldsymbol{\theta}}_k - \boldsymbol{\theta}^*||^2] = \mathbb{E}[||\hat{\boldsymbol{\theta}}'_k - \boldsymbol{\theta}^*||^2] \implies \frac{\operatorname{tr}(\boldsymbol{\Sigma})}{k^\alpha} = \boldsymbol{w}^T \boldsymbol{w} + \frac{\operatorname{tr}(\boldsymbol{\Sigma}')}{k^\alpha}. \tag{4.6}$$

As shown in Figure 4.1, when the calculated (via (4.6)) intersection point of the two graphs occurs at a small value of $k$, the calculated intersection from using the asymptotic normality result is not as accurate as when the intersection occurs at a large $k$. We can see this from the fact that the vertical line and the intersection of the blue and the orange curves do not overlap well when the true intersection manifests itself at a small $k$. The relatively poor-fit is because the normality, as an asymptotic result is not well established in small values of $k$. Regardless, we can see that our strategy of approximately calculating the point up until which the biased gradient estimator outperforms the unbiased one using the asymptotic normality of the iterates seems to be reasonable and effective. Note that even when the intersection occurs at a small $k$, the calculated intersection (i.e. vertical line) can be used as a reasonable estimate of the true intersection.

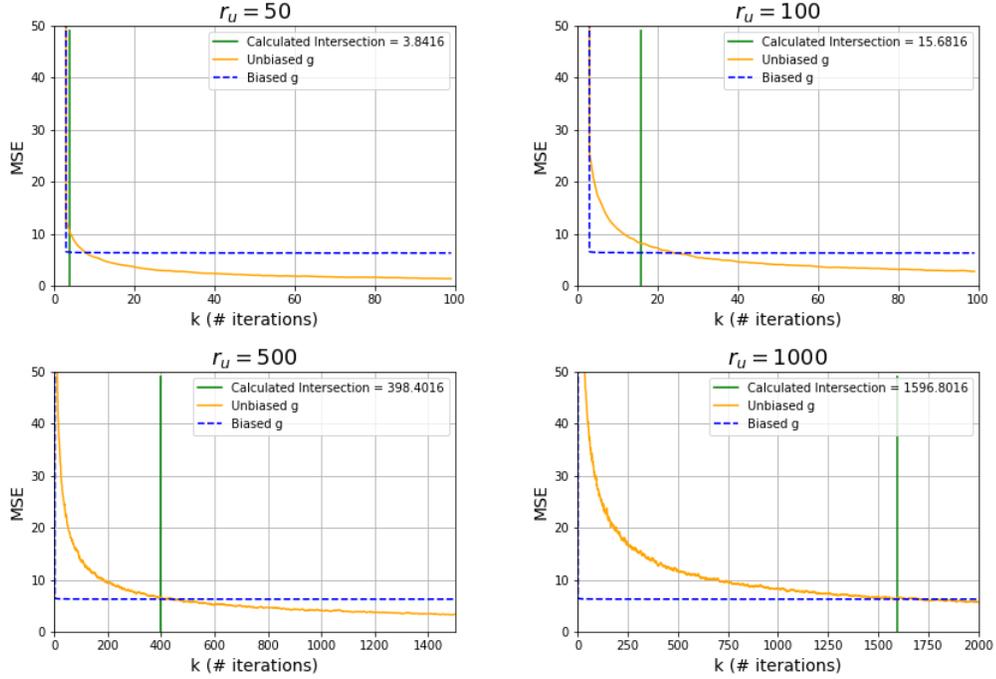

**Figure 4.1:** An illustration of how the MSE vs. iteration count of the biased and the unbiased SGD varies under different noise levels. We also show how accurate our theoretical calculation of the intersection point (green, vertical line) of the two curves is. In the plots, the blue (dashed) and the orange (solid) curves represent the MSE of the iterates generated by using biased and unbiased gradient estimators respectively.

## 4.2 SGD vs. SW-SGD Experiment

In this example, we analyze the effectiveness of the SW-SGD algorithm and show its superiority over the vanilla SGD in terms of the MSE of their iterates. To do so, we consider the skewed quartic loss function (from page 168 of (Spall, 2003)) given below as:

$$L(\boldsymbol{\theta}) = \sum_{i=1}^{p}(\boldsymbol{B}\boldsymbol{\theta})_i^2 + 0.1\sum_{i=1}^{p}(\boldsymbol{B}\boldsymbol{\theta})_i^3 + 0.01\sum_{i=1}^{p}(\boldsymbol{B}\boldsymbol{\theta})_i^4, \quad (4.7)$$

where $\boldsymbol{\theta} \in \mathbb{R}^p$, $\boldsymbol{B} \in \mathbb{R}^{p \times p}$ is such that $p\boldsymbol{B}$ is an upper triangular matrix of ones and $(\boldsymbol{B\theta})_i$ is the $i^{\text{th}}$ element of the vector $\boldsymbol{B\theta}$.

We let $p = 3$ and construct the noisy loss function as $Q(\boldsymbol{\theta}, V) = L(\boldsymbol{\theta}) \cdot V$ where $V \sim N(1, r)$. We run the experiment four times where we vary $r \in \{1, 2, 4, 8\}$. In all four of these set-ups, we let $\hat{\boldsymbol{\theta}}_0 = [10, 10, 10]^T$. Let us continue with calculating the SGD and SW gradient estimators. We calculated the $m^{\text{th}}$ element of the SGD gradient estimator to be the following:

$$(\hat{\boldsymbol{g}}(\hat{\boldsymbol{\theta}}_k))_m = \left[ 2 \sum_{i=1}^{p} \boldsymbol{B}^{(i,m)} \left( \sum_{j=1}^{p} \boldsymbol{B}^{(i,j)} \hat{\boldsymbol{\theta}}_k^{(j)} \right) + 0.3 \sum_{i=1}^{p} \boldsymbol{B}^{(i,m)} \left( \sum_{j=1}^{p} \boldsymbol{B}^{(i,j)} \hat{\boldsymbol{\theta}}_k^{(j)} \right)^2 \right.$$

$$\left. + 0.04 \sum_{i=1}^{p} \boldsymbol{B}^{(i,m)} \left( \sum_{j=1}^{p} \boldsymbol{B}^{(i,j)} \hat{\boldsymbol{\theta}}_k^{(j)} \right)^3 \right] V \qquad (4.8)$$

where $\boldsymbol{B}^{(i,j)}$ denotes the element in the $i^{\text{th}}$ row and the $j^{\text{th}}$ column of $\boldsymbol{B}$ and $\hat{\boldsymbol{\theta}}_k^{(j)}$ denotes the $j^{\text{th}}$ element of $\hat{\boldsymbol{\theta}}_k$. The SW gradient estimator is defined as in (3.3), where the $\hat{\boldsymbol{g}}(\hat{\boldsymbol{\theta}}_k)$ and $\hat{\boldsymbol{g}}(\hat{\boldsymbol{\theta}}_{k-1})$ terms that appear in (3.3) are computed using (4.8).

After running the experiment for four times with different values of $r$, we obtain the plots in Figure 4.2. Similar to the previous numerical study, we plotted these curves using empirical MSE values that were calculated over 1000 runs. First of all, as we can see in all four plots, the SW-SGD algorithm converges as the blue (dashed) curves are converging to an MSE of 0. This numerical finding supports the proof of convergence that was presented in Chapter 3 for the SW-SGD algorithm. Furthermore, we observe that for all values of $r$ (i.e. for different noise levels), the SW-SGD algorithm is outperforming the SGD algorithm as the blue (dashed) curves are below the orange (solid) curves in all four plots.

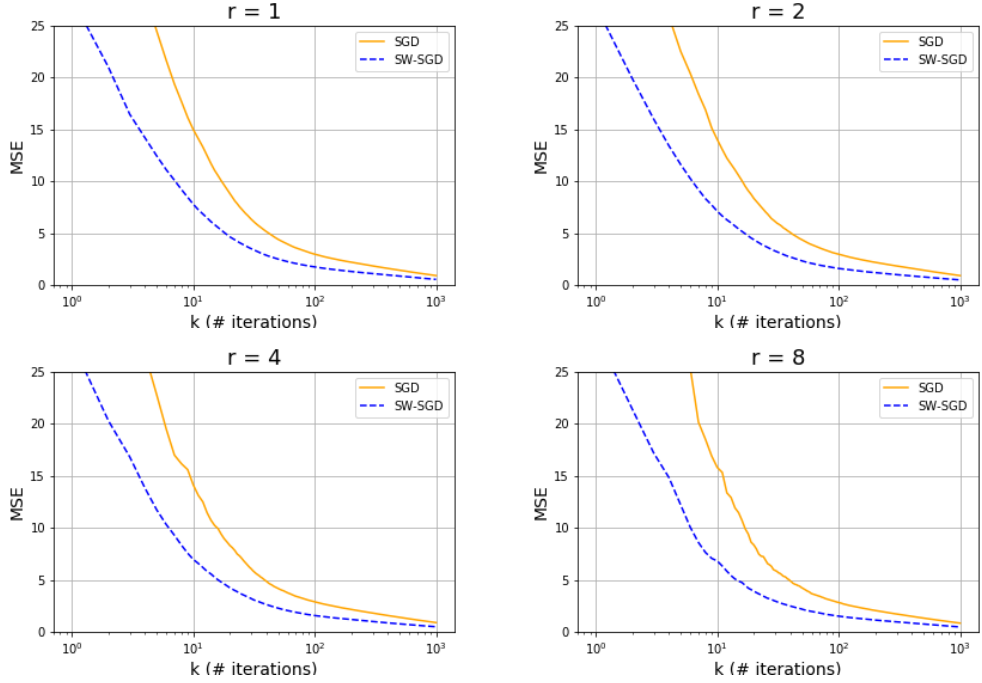

**Figure 4.2:** An illustration of the MSE vs. iteration count of the $\hat{\theta}_k$ iterates obtained by using the SGD and SW-SGD update under different noise levels. In the plots, the orange (solid) and the blue (dashed) curves represent the empirical MSE of the iterates generated from the SGD and SW-SGD updates respectively.

We see that for low values of $r$, the performances (in an MSE sense) of the SGD and SW-SGD algorithms are similar, with SW-SGD leading with a small margin. As $r$ increases, we observe that the gap between the blue and the orange curves widen in favor of the SW-SGD algorithm. This happens because an increase in the variance of $V$ affects the SW-SGD algorithm less due to its variance reduction property. Even though the SW gradient estimators are slightly biased, the fact that they have reduced variances compared to the SGD gradient estimators makes them more effective in this problem.

Furthermore, we can observe that the stability of the SGD algorithm is significantly affected as $r$ increases relative to the SW-SGD algorithm. As $r$ increases the amount of fluctuations in the orange curves increase whereas the blue curves are still relatively stable. So overall this numerical study with the insights that we gathered provide an empirical evidence to the effectiveness of the SW-SGD algorithm.

# Chapter 5

# Conclusion and Future Directions

## 5.1 Conclusion

This paper highlights the potential usefulness of biased stochastic gradient algorithms with reduced variance. We clearly see that there exists a tradeoff between the bias and the variance of the gradient estimators. Using the asymptotic normality result of Fabian, we were able to provide a general framework to precisely characterize the MSE of the iterates of stochastic gradient algorithms.

Even though Fabian's result is asymptotic, we were able to show that in practice the normality manifests itself quite rapidly and is almost fully established for large $k$. Regardless, it is important to note that this is a significant limitation of this paper and the normality that we use for the iterates is merely an assumption that relies on the asymptotic result.

Our main contribution in this paper was the SW-SGD algorithm where we were able to give a proof of convergence and establish asymptotic normality.

Furthermore, through our numerical studies, we showed that SW-SGD incurred a lower MSE overall than the SGD algorithm and that it is more robust to increase in variance of the randomness in stochastic optimization problems (see the numerical study in Section 4.2).

## 5.2 Future Directions

We believe that there are a few natural paths to pursue. One of these paths would be to explore how the SW-SGD performs for different values of the window size (i.e. $t$). Perhaps it might be possible to express the MSE of the SW-SGD algorithm in terms of the window size $t$ and then find the optimal window size by minimizing the MSE with respect to $t$. However, it is important to point out that with the assumptions and the analysis provided in Section 3.2, the proof of convergence of our SW-SGD algorithm cannot be readily generalized to $t > 2$. So before parameterizing and minimizing the asymptotic MSE with respect to $t$, one would have to obtain a proof of convergence (in an almost sure sense).

It would also be interesting to analyze the MSE of the already existing or novel stochastic gradient algorithms (that exploit biased gradients) using the principles presented in this paper that rely on Fabian's work. A bias-variance tradeoff analysis similar to the one we adopted in this paper could also be adopted to gradient-free methods such as SPSA (Spall, 1992). Also, we could generalize the ideas presented in this paper to fit into the framework of feedback control in dynamical systems as in (Spall and Cristion, 1994). Having a sliding window estimator might help provide additional stability in a potentially unstable environment. In (Qian, 1999), the author also constructs momentum-based gradient estimator similar to our SW gradient estimator and then finds the upper and lower bounds for the coefficients of his update equation that would guarantee convergence. Assuming that a natural follow-up on this introduction of SW gradient estimator would be to explore more generalized and parameterized versions of it, the ideas in (Qian, 1999) could shed light towards a potential path. For example, if we decide to analyze the performance of the SW-SGD algorithm for larger window sizes, it might be a good idea to consider weighting these gradients such that the older ones contribute less information. The optimal weighing of the gradients could be found using the strategies in (Qian, 1999).

As we have mentioned in the previous section, using an asymptotic result for finite analysis is an approximation, although the results are fairly accurate in the studies here. To make the approach more exact, one could try to construct an upper bound on the error that is incurred when normality is assumed for the finite iterates. This would basically require obtaining a result similar to the Bery-Esseen Theorem for the sample mean. Alternatively, the finite-sample analysis of SA in (Liu, Hou, and Spall, 2019) might be useful.

To extend the SW-SGD algorithm to more general cases, we could try to relax the assumption we have on the noise of our SO problem. In particular, we could try to obtain a proof of convergence for the SW-SGD algorithm without assuming that the variance of the noise sequence decays.

Overall, our analysis showed that biased gradient estimators can be beneficial over the unbiased gradient estimators under some conditions. Through our results, theoretical and numerical, we can infer that this tradeoff is worth pursuing and analyzing further in various other SO setups.